\documentclass[runningheads]{llncs}
\usepackage{graphicx}
\usepackage{cite}
\usepackage{amsmath}
\usepackage{amssymb}
\usepackage{hyperref}
\usepackage{multirow}
\usepackage{pifont}
\usepackage{graphicx}
\usepackage{wrapfig}
\graphicspath{{./images/}}
\usepackage{floatrow}
\newfloatcommand{capbtabbox}{table}[][\FBwidth]

\begin{document}
\title{CP-VoteNet: Contrastive Prototypical VoteNet for Few-Shot Point Cloud Object Detection}
\titlerunning{CP-VoteNet}
% If the paper title is too long for the running head, you can set
% an abbreviated paper title here
%
\author{Xuejing Li
%\inst{1}\orcidID{0000-0001-7145-7225} 
 \and
 Weijia Zhang
% \inst{1}\orcidID{1111-2222-3333-4444} 
 \and
 Chao Ma
% \inst{1}\orcidID{2222--3333-4444-5555}
}
\authorrunning{X. Li et al.}
% First names are abbreviated in the running head.
% If there are more than two authors, 'et al.' is used.
%
\institute{Shanghai Jiao Tong University\\
 \email{lixuejing, weijia.zhang, chaoma@sjtu.edu.cn}}
%\url{http://www.springer.com/gp/computer-science/lncs} \and
% ABC Institute, Rupert-Karls-University Heidelberg, Heidelberg, Germany\\
% \email{\{abc,lncs\}@uni-heidelberg.de}}
% }
%
\maketitle              % typeset the header of the contribution
\begin{abstract}
Few-shot point cloud 3D object detection (FS3D) aims to identify and localise objects of novel classes from point clouds, using knowledge learnt from annotated base classes and novel classes with very few annotations. 
Thus far, this challenging task has been approached using prototype learning, but the performance remains far from satisfactory. 
We find that in existing methods, the prototypes are only loosely constrained and lack of fine-grained awareness of the semantic and geometrical correlation embedded within the point cloud space. 
To mitigate these issues, we propose to leverage the inherent contrastive relationship within the semantic and geometrical subspaces to learn more refined and generalisable prototypical representations.
To this end, we first introduce contrastive semantics mining, which enables the network to extract discriminative categorical features by constructing positive and negative pairs within training batches. 
Meanwhile, since point features representing local patterns can be clustered into geometric components, we further propose to impose contrastive relationship at the primitive level. Through refined primitive geometric structures, the transferability of feature encoding from base to novel classes is significantly enhanced.
The above designs and insights lead to our novel Contrastive Prototypical VoteNet (CP-VoteNet). 
Extensive experiments on two FS3D benchmarks FS-ScanNet and FS-SUNRGBD demonstrate that CP-VoteNet surpasses current state-of-the-art methods by considerable margins across different FS3D settings. Further ablation studies  conducted corroborate the rationale and effectiveness of our designs.

\keywords{few-shot point cloud object detection \and contrastive learning \and prototype learning.}
\end{abstract}
\section{Introduction}
% What is point cloud 3OD
Point cloud 3D object detection aims to identify and locate objects in 3D space from point cloud scenes. It finds significant application across a spectrum of 3D scene understanding tasks, such as autonomous driving, robotics, and augmented reality. 
% Problem with point cloud 3OD
The recent success of deep-learning-driven point cloud 3D object detectors has, however, largely hinged on fully-supervised training with enormous amounts of labelled training data whose manual annotation is costly, time-consuming, and labor-intensive. 
% FSL helps
To ease the burden of data annotation, few-shot learning (FSL)~\cite{metalearning, prototypicalFSL, matchingFSL, relationFSL, metafsl, sgdfsl} has been proposed as an appealing data-efficient machine learning paradigm for point cloud learning, alongside semi-~\cite{SESS, 3DIoUMatch, DDS3D, ODM3D} or weakly-supervised~\cite{HybridCR, MIT, WS3D} set-ups. 
% What does FSL do
FSL enables the network to execute scene understanding tasks for novel classes using knowledge learnt from base classes with abundant annotations and novel classes with very few annotated samples.
It has shown success in a plethora of computer vision tasks, including image classification~\cite{metalearning, prototypicalFSL, matchingFSL, relationFSL}, semantic segmentation~\cite{panetFSS, priorFSS, adaptiveFSS, partFSS}, and 2D object detection~\cite{frustratingfsod, distillfsod, transformerfsod}.

Very recently, few-shot learning has also been explored for the task of 3D object detection from point clouds. These preliminary investigations on few-shot point cloud 3D object detection (FS3D)~\cite{MetaDet, GeneralizedFS3D, zhao2022prototypical, PrototypicalVAE} mostly opt for a prototype learning approach.
Specifically, they extract various prototypical representations (also referred to as ``prototypes''), abstractions of fundamental semantic or geometric properties, from labelled support samples to guide object detection in unlabelled query scenes. These prototypical networks posit that semantic categories and geometric structures have prototype representations, which can offer guidance for object detection and knowledge that is transferable to novel classes. 

Nevertheless, upon careful investigation we find that these early methods extract prototypes in a rather simplistic and unconstrained manner. The resultant prototypes lack awareness of fine-grained semantics and geometric component, as well as acute perception of the correlation amongst these prototypes. These have led to sub-optimal performance when transferring to novel classes. Inspired by the success of contrastive learning as a powerful representation learning methods \cite{simCLRv1, MoCo, InstDisc, InvaSpread} and their application in other few-shot or point cloud learning tasks~\cite{FSCCL, FSE, SCLFSC, RobustSeg, 3DCoCo, ProtoTransfer, CrossPoint}, in this paper we propose to leverage contrastive learning to learn more refined, semantic- and geometry-aware prototypes for FS3D.

\begin{figure}[t]
\centering\includegraphics[width=\textwidth]{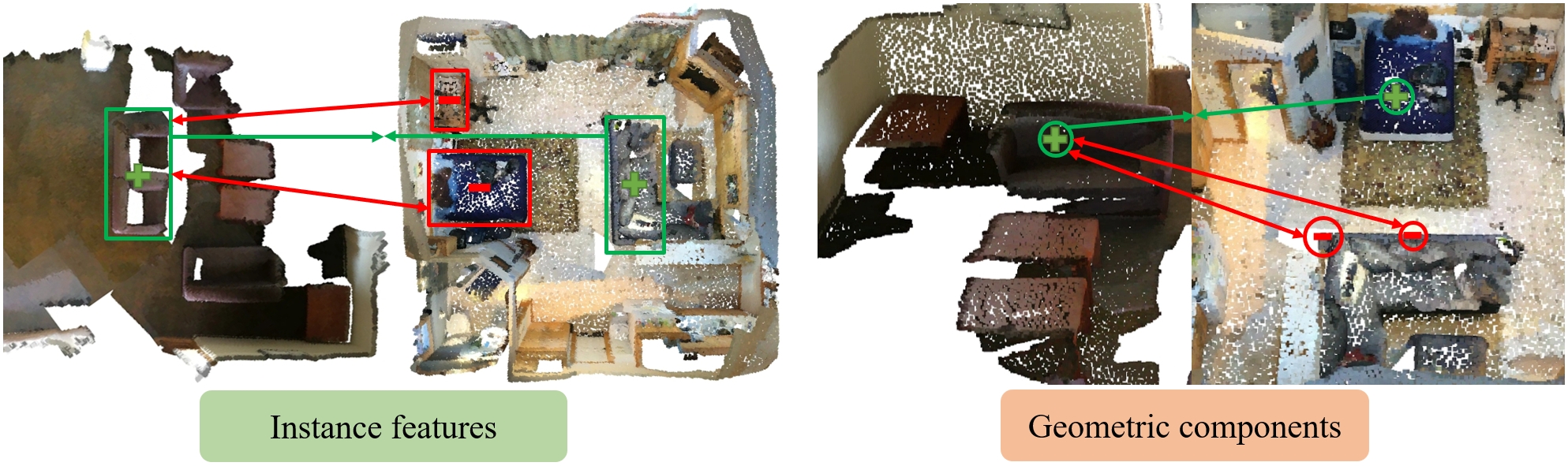}
\caption{Contrastive learning at the semantic level (\textbf{left}) and geometric level (\textbf{right}). Semantic contrastive learning requires that the instance features of positive pairs belonging to the same category be similar, and those of negative pairs belonging to different categories be dissimilar. Primitive contrastive learning demands that features of the same geometric components (i.e., faces within green circles) be similar, and those of different geometric components (i.e., edges and corners within red circles) be dissimilar.} \label{fig:attraction}
\end{figure}

In light of the above analysis, we design an tailored to the FS3D task, named Contrastive Prototypical VoteNet (CP-VoteNet). As shown in Fig. \ref{fig:attraction}, our method consists of two parts: semantic contrastive learning (SCL) and primitive contrastive learning (PCL). The former operates on the support instances, necessitating that instance features within the same semantic category exhibit similarity, while actively distancing features of objects across disparate categories. The latter applies to all inputs, demanding that features of points belonging to the same primitive geometric structure should be similar to its geometric prototype, and divergent from features not belonging to this structure, regardless of whether these points fall within the same semantic category or not. By appropriately balancing the contributions of these two contrastive learning strategies, we synergistically enhance both the discriminative and generalisation abilities of the network. 
It is worth noting that, in contrast to strategies that require data augmentation to generate positive and negative sample pairs, our sample pairs are contructed within batches, eliminating the need for a significant increase in computational load and computing time.

We conduct experiments to evaluate the effectiveness of our proposed method on two FS3D benchmarks: FS-SUNRGBD and FS-ScanNet, following\cite{zhao2022prototypical, PrototypicalVAE}. The results reveal that our method outperforms existing methods by considerable margins across different FS3D settings. We also perform ablation studies to validate the rationality and effectiveness of our designs.

Our contributions are summarised as follows:

\begin{itemize}
    \item [$\bullet$] We design semantic contrastive learning to explore the intrinsic feature space of semantic prototypes, which refines the network's semantic feature extraction and discrimination by imposing constraints on features at the semantic level.
    \item [$\bullet$] We introduce contrastive regularisation to primitive feature learning for a more constrained geometric feature space. This results in enhanced structural awareness and nuanced local geometry information within learnt prototypes, boosting their generalisation on novel categories.
    \item [$\bullet$]We present the concise and effective CP-VoteNet, which achieves state-of-the-art performance in most task settings. To our best knowledge, this is the first time elements of contrastive learning are introduced into FS3D.
\end{itemize}

\section{Related Work}

\subsubsection{Few-Shot 2D Object Detection}
Few-shot 2D object detection (FS2D)\cite{kangfsod, fanfsod, multifsod, distillfsod, transformerfsod, frustratingfsod} methods can be categorised into two types: two-branch approaches and single-branch approaches. 
Typically, two-branch methods\cite{kangfsod, fanfsod, frustratingfsod} take a set of both query and support images as input and  focus on designing the interaction between a query branch and a support branch. Relevant features of query images are extracted and refined under the guidance of annotated support images. Single-branch methods\cite{frustratingfsod, distillfsod} are usually first trained on base classes with a large number of annotated samples to equip the network with basic detection capabilities. Once pre-trained, they are then fine-tuned on a few annotated novel samples to improve the network's adaptability to novel classes. Overall, current FS3D methods predominantly draw on the idea of using support to guide query and conduct two-stage training comprising pre-training and fine-tuning.

\subsubsection{Few-Shot 3D Object Detection} 
Few-shot 3D object detection from point clouds (FS3D) has been a less visited problem.
Existing FS3D approaches~\cite{MetaDet, GeneralizedFS3D, PrototypicalVAE, zhao2022prototypical} are predominantly grounded in prototype learning. Generalised FS3D\cite{GeneralizedFS3D} introduces and fine-tunes additional detection heads for novel classes, enhancing the detection capabilities for these classes while maintaining the performance for base classes, and employs a sample adaptive balance loss to address the issue of class imbalance. MetaDet3D\cite{MetaDet} extracts class-specific re-weighting vectors from support instances to guide voting and the generation of object proposals. Besides semantic prototypes, Prototypical VoteNet\cite{zhao2022prototypical} utilises a memory bank to record class-agnostic geometric prototypes, which are later used for local feature enrichment. The current top-performer, P-VAE~\cite{PrototypicalVAE} incorporates an auxiliary task beyond the detection task that leverages prototypes to aid in scene and object reconstruction, thereby regularising prototype learning.

\subsubsection{Contrastive Learning}
Contrastive learning\cite{InstDisc, InvaSpread, MoCo, simCLRv1, HybridCR} is initially proposed as a learning paradigm for effective representation learning from unlabelled data. It demands the model to recognise the invariance and distinctiveness in data by constructing positive and negative sample pairs. 
In the domain of point cloud scene understanding, HybridCR~\cite{HybridCR} proposes two contrastive learning strategies with a novel dynamic point cloud augmentor for weakly-supervised point cloud segmentation. 
Contrastive learning has expanded from unsupervised learning to few-shot learning~\cite{FSCCL, SCLFSC, FSE}. 
ContrastBoundary~\cite{ContrastBoundary} applies contrastive learning at the boundaries of different instances, sharpening their edges in the FS3D-Seg (few-shot point cloud segmentation) task. 
R3DFSSeg\cite{RobustSeg} uses class-wise contrastive learning on local feature components in the field of FS3D-Seg to enhance the differences amongst support shots of different categories and enhance the similarity of the same category. Contrastive learning has been proven widely effective in FS3D-Seg, but its effectiveness in the FS3D task remains indeterminate. 
Due to the inherent discrepancy in the granularity of learnt representations between FS3D and FS3D-Seg, applying contrastive learning in the former requires tailored design.

\section{Proposed Method}
This section describes in detail the technical aspects of the proposed CP-VoteNet. To set its stage, we first present the problem definition of the FS3D task in section~\ref{section:definition}. Next, we present a high-level outline of the CP-VoteNet framework in section~\ref{section:protovotenet}, followed by detailed explanation on our designs in sections~\ref{section:semantic} and \ref{section:primitive}. Finally, we formulate the overall training objective of CP-VoteNet in ~\ref{section:loss}.

\subsection{Problem Definition} \label{section:definition}
Following the dataset settings of FS3D \cite{MetaDet, zhao2022prototypical, PrototypicalVAE}, the entire dataset $\mathbb{D}$ is divided into two parts: $\mathbb{D}_{base}$ and $\mathbb{D}_{novel}$, such that $\mathbb{D}_{base} \bigcup \mathbb{D}_{novel} = \mathbb{D}$ and $\mathbb{D}_{base} \bigcap \mathbb{D}_{novel} = \O $. $\mathbb{D}_{base}$ contains base classes where all instances have detailed annotations, including bounding boxes and category information, while in $\mathbb{D}_{novel}$, only a few instances in each category have such annotations. 
The goal of FS3D is to detect novel classes through knowledge and capabilities acquired from the full annotations in $\mathbb{D}_{base}$ and limited annotations in $\mathbb{D}_{novel}$.

For an \textit{N-way K-shot} FS3D task, there are \textit{N} categories to be detected in $\mathbb{D}_{novel}$, where each class has \textit{K} objects with detailed annotations, which are referred to as support samples, while all other objects are unannotated, which are referred to as query samples. During the testing phase, point cloud scene $Q$ containing query samples and their corresponding $N\times K$ support samples $\mathbb{S}$ are fed as input into the network. Then, query samples are detected under the guidance of support ones. During the training phase, following the meta-leaning paradigm~\cite{metalearning}, a batch of input $I$ is also organised into multiple sub-tasks called episodes in the form of $I=\{\mathbb{S}^b,Q^b\}_{b=1}^B$, where $B$ is the batch size and $\mathbb{S}=\{S_{n,k}\}_{n=1,k=1}^{N,K}$. In each iteration, the model learns how to detect query objects leveraging knowledge learnt from $\mathbb{S}$ through these episodes.

\begin{figure}[t]
\includegraphics[width=\textwidth]{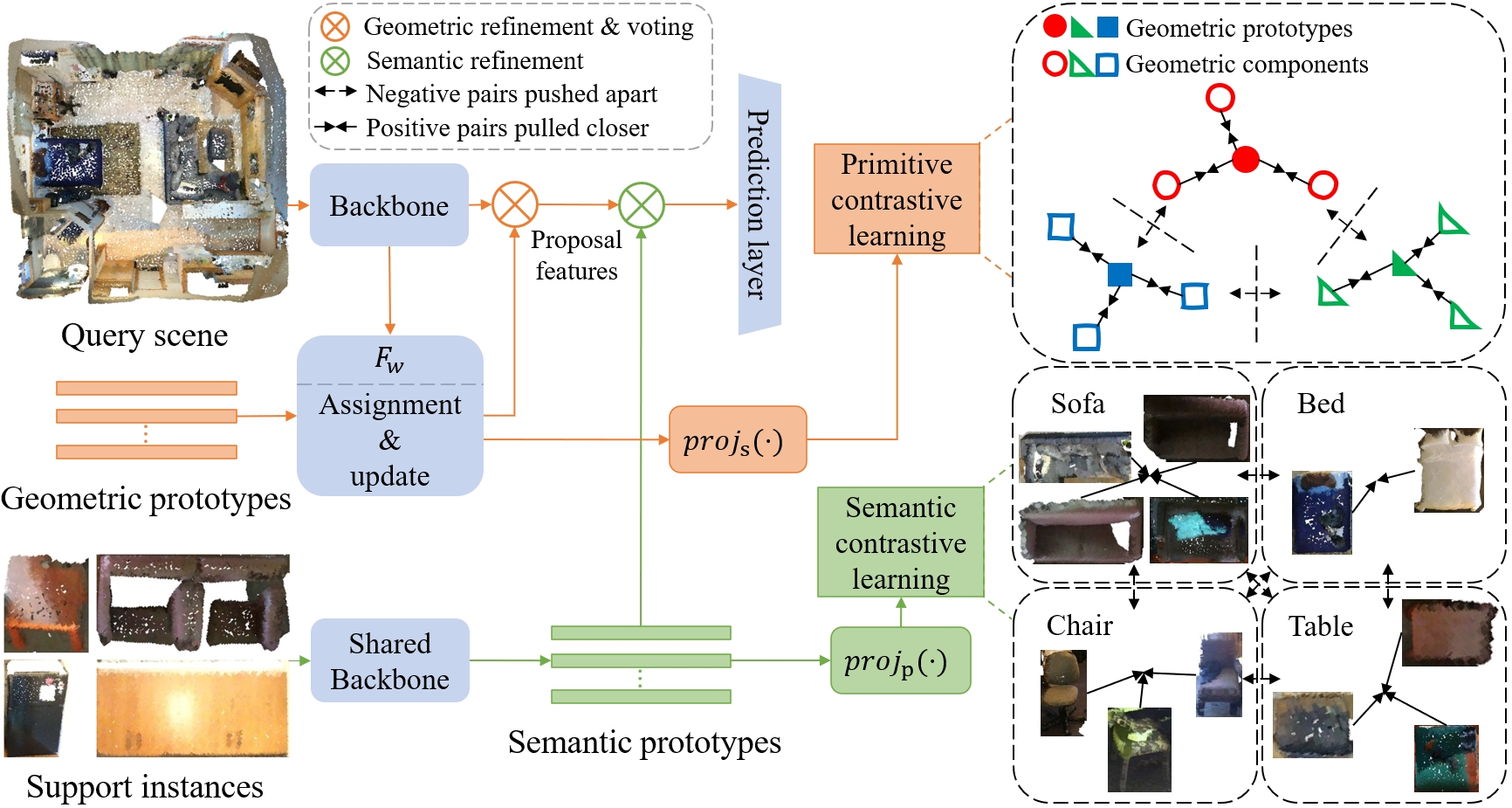}
\caption{The overall framework of the proposed CP-VoteNet. Positive and negative pairs are constructed within a minibatch for contrastive learning. Features $F_w$ assigned to different geometric prototypes are considered as different geometric components, which, after passing through a projection layer $proj_{\text{s}}(\cdot)$, engage in primitive contrastive learning. The instance features of the support instances, serving as semantic prototypes and passing through a projection layer $proj_{\text{p}}(\cdot)$, undergo semantic contrastive learning.} \label{fig:framework}
\end{figure}

\subsection{Overall Framework} \label{section:protovotenet}
With Prototypical VoteNet~\cite{zhao2022prototypical} as our baseline and PointNet++\cite{PointNet} its point feature extractor, we first outline the overall framework of our method.

Given an input point cloud scene $S_i$ with $N$ points, the 3D backbone will extract $M$ feature points with $d$-dimensional features and $3$-dimensional positional coordinates as seed points. We denote these seed features extracted by backbone as $F_i=f_i \oplus d_i$, where $f_i\in \mathbb{R}^{M\times d}$, $d_i\in \mathbb{R}^{M\times 3}$, and $\oplus$ represents the concatenation operation.

In the pretraining stage, a randomly initialised memory bank is used to record the geometric components of $S_i$, which are later referred to as geometric prototypes $G=\{g_w\}^W_{w=1}\in \mathbb{R}^{W\times d}$. $W$ is the manually preset number of components. Specifically, the point features $f_i$ belonging to foreground will be assigned to $G$ in order of similarity and update $G$ following Eq. (\ref{eq:GW}).

\begin{equation}
g_w\leftarrow \gamma g_w + (1-\gamma)f_w,\gamma\in (0,1), f_w=Average(\{f_i\}_w) \label{eq:GW}
\end{equation}
where $\{f_i\}_w$ are the point features assigned to $g_w$, and $\gamma$ is the momentum coefficient facilitating the update of $f_w$ into geometric prototypes in a momentum manner.

Since each foreground $f_i$ is assigned a unique $g_w$, we can obtain the primitive pseudo-labels for foreground points. In this way, primitive contrastive learning can be performed by constructing positive and negative sample pairs based on these pseudo labels, which will be explained in detail in section \ref{section:primitive}.

In the following module, seed features are refined with geometric prototypes through a cross-attention module and residual connection. Then, in the query branch, $F_i$ are voted and clustered to form features of region proposals, while in the support branch, $f_i$ are pooled to generate class-specific semantic prototypes, where semantic contrastive learning are performed and will be further explained in section \ref{section:semantic}

Finally, proposal features are refined with semantic prototypes. The detection results of bounding boxes, objectiveness scores, and classification logits for each object proposal are generated by the subsequent prediction layer.

\subsection{Semantic Contrastive Learning}\label{section:semantic}
In this section, we present the design of our semantic contrastive learning for FS3D tasks. Compared to point cloud segmentation, the feature representation in point cloud detection tends to be local patterns rather than just point features. So for FS3D, the granularity of point cloud features is coarser than that of FS3D-Seg, and thus the contrastive learning using fine-grained point features is relatively difficult to implement. Moreover, the feature in FS3D not only reflects the characteristics of its category, but also the geometric elements at its location, such as angles, edges, faces, and other geometric components that make up a three-dimensional object. Based on the above analysis, it is obvious that we cannot ignore the geometric information covered by seed features. Roughly converging features of seed points of the same class and distancing those of different classes can cause semantic-level overfitting and the loss of geometric information possessed by different geometric components. Therefore, we propose a prototype-level semantic contrastive learning strategy that takes effect on support prototypes, avoiding imposing mandatory constraints on specific seed points. The following is a detailed description.

According to the training strategy described in section \ref{section:definition}, base classes are divided into small detection tasks as training data for the network. A batch of size $B$ has objects belonging to $N$ categories to be detected, and each category should contain $K$ annotated instances. So, in the input training data, one query scene corresponds to $N\times K$ labelled support instances.

In the feature extraction process of support instances, the backbone extracts seed features $\{f_i\}_n$, where $i$ refers to seed point number, and $n$ refers to category. An instance feature $f_n=Average\{f_i\}_n$ is later used as a class prototype.

Let support instance features of the $n$-th class corresponding to the $b$-th query scene in a batch be $F_n^b=\{f_k\}_n^b, b\in[1,B], k\in[1,K], n\in[1,N]$. For the sake of computational simplicity and to prevent overfitting at the semantic level, take general support prototype of the n-th category $P_n^b=\frac{1}{|F_n^b|}\sum\limits_{k}{\{f_k\}_n^b}$ for subsequent process of semantic contrastive learning.

Inspired by SimCLR~\cite{simCLRv1}, we incorporate a projection layer $proj_{\text{p}}(\cdot)$ to project $P$ onto a different feature space for similarity computation. Computing similarity in this new space refrains the calculated loss from directly updating the original feature via backpropagation, which potentially leads to loss of salient feature information. 
$proj_{\text{p}}(\cdot)$ is simply implemented as an MLP layer and is employed during training only. The feature $P$ used to calculate similarity is generated following $P \leftarrow proj_{\text{p}}(P)$.

For the input of one batch, similarity between class $n$ in task $b$ and class $m$ in other tasks is expressed as
\begin{equation}
\text{sim}(b,n,m)=
\begin{cases}
\frac{1}{B}\sum^B\limits_{i=1}\text{sim}(P_n^b,P_m^i), & n\neq m \\
\frac{1}{B-1}\sum^B\limits_{i=1,i\neq b}\text{sim}(P_n^b,P_m^i), & n=m
\end{cases}
\end{equation}
where $\text{sim}(P_n^b,P_m^i)$ is calculated by vector dot product. Our semantic contrastive loss is formulated on the basis of InfoNCE loss\cite{infoNCE, MoCo} and is given by.
\begin{equation}
    L_{\text{semcl}}=-\frac{1}{B}\frac{1}{N}\sum\limits_{b=1}^B\sum\limits_{n=1}^Nlog
    \frac{exp(\text{sim}(b,n,n)/\tau)}{\sum_{m=1}^Nexp(\text{sim}(b,n,m)/\tau)}
\end{equation}
where $\tau$ is a temperature parameter to adjust the smoothness of the similarity distribution. 

\subsection{Primitive Contrastive Learning}\label{section:primitive}
By narrowing the proximity between similar samples and expanding the distance between dissimilar ones within the data space, contrastive learning fundamentally enhances the clustering efficacy in said space. Consequently, contrastive learning can be applied to any data space where viable clustering exists.

The geometric prototypes formed through similarity-based assignment and momentum updates, as described in section \ref{section:protovotenet}, can also be viewed as a process of clustering foreground seed features. Since the features of each foreground seed point are assigned to only one geometric prototype, this assignment allows for the labeling of each seed point with a geometric pseudo label which is, in essence, the index of the geometric prototype. Based on these labels, we can construct positive and negative pairs within the geometric feature space. Since the geometric prototype before updating does not contain the newly assigned seed feature, we regard the prototype and the feature assigned to it as a positive pair, and the feature assigned to other prototypes as a negative pair. In this way, the foreground seed features extracted by the network exhibit more distinct characteristics of the geometric components they represent.

Given geometric prototypes $G=\{g_w\}^W_{w=1}$, point features assigned to $g_w$ are $F_w=\{f_i\}_w$. For the convenience of calculation and to avoid geometric-level overfitting, $M_w=\frac{1}{|F_w|}\sum\limits_{i}\{f_i\}_w$ is used to calculate the primitive contrastive loss.

Similar to section \ref{section:semantic}, a projection layer $proj_{\text{p}}(\cdot)$ is added between $M$ and similarity calculation, and similarity is calculated by matrix dot multiplication. The feature $M$ used for primitive contrastive learning is $M \leftarrow proj_{\text{p}}(M)$.In this way, the primitive contrastive loss can be expressed as
\begin{equation}
    L_{\text{primcl}}=-\frac{1}{W}{\sum\limits_{w=1}^{W}log\frac{{exp(\text{sim}(M_w,g_w^*)/\tau)}}{\sum{_{j=1}^Wexp(\text{sim}(M_j,g_w^*)/\tau)}}}
\end{equation}
where * indicates the detachment of the gradients for the variable to prevent update on it during backward propagation.

\subsection{Loss Function} \label{section:loss}
To maintain both the generalisation ability across categories and the discriminative ability within categories, we combine the objective functions described in sections~\ref{section:semantic} and~\ref{section:primitive} into the following overall training objective:
\begin{equation}
L=L_{\text{det}}+\lambda_1L_{\text{semcl}}+\lambda_2L_{\text{primcl}}
\end{equation}
where $L_{\text{det}}$ is the original loss function proposed in Prototypical VoteNet~\cite{zhao2022prototypical}, which includes losses of classification, bounding box regression, objectness prediction, and offset voting~\cite{PointNet}; $\lambda_1$ and $\lambda_2$ are coefficients to balance the contribution of two contrastive losses to ensure that the network still focuses on optimising $L_{\text{det}}$ as its main training objective. Moreover, by adjusting $\lambda_1$ and $\lambda_2$, we prevent the network from overfitting on the semantic information of base classes, which will occur when the impact of $L_{\text{semcl}}$ far exceeds that of $L_{\text{primcl}}$, as well as from losing semantic discrimination ability, which can occur when the situation is reversed. The impact of imbalanced $L_{\text{semcl}}$ and $L_{\text{primcl}}$ will be quantitatively demonstrated through ablation experiments in section \ref{section:ablation}.

\section{Experiments}

\subsection{Experiment Setup}
\subsubsection{Datasets} 
Following prior works~\cite{prototypicalFSL, PrototypicalVAE, MetaDet}, our method is trained and evaluated on FS-ScanNet and FS-SUNRGBD~\cite{zhao2022prototypical}---two derivative datasets from ScanNet~\cite{scannet} and SUNRGBD~\cite{SUNRGBD}, respectively, tailored to few-shot 3D indoor scene understanding. 
 
\textbf{FS-ScanNet} is composed of 1513 point clouds in 18 annotated semantic categories. 6 out of the annotated categories are randomly selected as novel classes, while the remaining categories are considered base classes. FS-ScanNet utilises two sets of divisions, namely Split-1 and Split-2, for base and novel classes. Each novel class contains $k$ labeled instances, where $k$ equals to 1, 3, and 5. 
\textbf{FS-SUNRGBD} consists of 5000 annotated RGB-D samples covering 10 categories. 4 classes are randomly selected as novel ones, keeping other classes as base ones. Only $k$ instances are annotated in each novel class, where $k$ equals to 1 to 5.

\subsubsection{Implementation Details} Our training and evaluation configurations are consistent with those of Prototypical VoteNet \cite{zhao2022prototypical}.
Specifically, we train CP-VoteNet with a batch size of 16, a maximum epoch number of 36 during pretraining, 5 during finetuning, a learning rate of 0.008, 128 geometric prototypes, momentum update $\gamma$ of 0.999, and weight decay of 0.01 for the AdamW optimiser. Specifically, $\lambda_1$ and $\lambda_2$ are set to 0.1. In the projection layer, the feature length is reduced to half of its original size, from 256 to 128. For the calculation of contrastive loss, $\tau$ is set to 0.2. For evaluation, we follow the standard 3D object detection evaluation protocol to report the mean average precision AP$_{25}$ and AP$_{50}$, at IoU thresholds of 0.25 and 0.50, respectively.

\begin{table}[t]
\centering  \tabcolsep=0.02cm
\caption{Results on \textbf{FS-ScanNet} and comparison against existing works of FS3D.} \label{tab:ScanNet}
\resizebox{\textwidth}{!}{
\begin{tabular}{l|cc|cc|cc|cc|cc|cc}
\hline
    \multirow{3}{*}{\parbox{3.5cm}{\centering \textbf{Method}}} 
    & \multicolumn{6}{c|}{\bfseries Novel Split 1} & \multicolumn{6}{c}{\bfseries Novel Split 2} \\
    \cline{2-13}
    & \multicolumn{2}{c|}{\bfseries 1-shot} & \multicolumn{2}{c|}{\bfseries 3-shot} & \multicolumn{2}{c|}{\bfseries 5-shot} & \multicolumn{2}{c|}{\bfseries 1-shot} & \multicolumn{2}{c|}{\bfseries 3-shot} & \multicolumn{2}{c}{\bfseries 5-shot} \\
    \cline{2-13}
    & AP$_{25}$ & AP$_{50}$ & AP$_{25}$ & AP$_{50}$ & AP$_{25}$ & AP$_{50}$ & AP$_{25}$ & AP$_{50}$ & AP$_{25}$ & AP$_{50}$ & AP$_{25}$ & AP$_{50}$ \\
\hline
Prototypical VoteNet~\cite{zhao2022prototypical} & 15.34 & 8.25 & 31.25 & 16.01 & 32.25 & 19.52 & 11.01 & 2.21 & 21.14 & 8.39 & 28.52 & 12.35\\
\hline
MetaDet3D\cite{MetaDet} & 10.28 & 4.03 & 23.42 & 10.64 & 25.65 & 13.88 & 5.21 & 1.32 & 15.44 & 4.37 & 22.13 & 7.09\\
Generalized FS3D\cite{GeneralizedFS3D} & 12.03 & 8.19 & 24.90 & 10.26 & 29.29 & 16.67 & 9.19 & 1.87 & 19.41 & 6.80 & 25.18 & 12.74 \\
P-VAE\cite{PrototypicalVAE} & 16.00 & 10.22 & 31.60 & \textbf{19.37} & 32.84 & 22.39 & 12.66 & 4.15 & 21.27 & 10.09 & 31.70 & 14.43\\
\hline
\bfseries CP-VoteNet & \bfseries16.48 & \bfseries12.48 & \bfseries34.13 & 18.83 & \bfseries 36.61 & \bfseries 24.36 & \bfseries15.41 & \bfseries5.03 & \bfseries26.60 & \bfseries10.83 & \bfseries35.02 & \bfseries16.07 \\
\hline
\end{tabular}
}
\end{table}

\begin{table}[t] \centering \scriptsize
\caption{Results on \textbf{FS-SUNRGBD} and comparison against existing works of FS3D.} \label{tab:SUNRGBD}
\resizebox{0.99\textwidth}{!}{
\begin{tabular}{l|cc|cc|cc|cc|cc}
\hline
    \multirow{2}{*}{\parbox{3cm}{\centering \textbf{Method}}} & \multicolumn{2}{c|}{\bfseries 1-shot} & \multicolumn{2}{c|}{\bfseries 2-shot} & \multicolumn{2}{c|}{\bfseries 3-shot} & \multicolumn{2}{c|}{\bfseries 4-shot} & \multicolumn{2}{c}{\bfseries 5-shot} \\
    \cline{2-11}
    & AP$_{25}$ & AP$_{50}$ & AP$_{25}$ & AP$_{50}$ & AP$_{25}$ & AP$_{50}$ & AP$_{25}$ & AP$_{50}$ & AP$_{25}$ & AP$_{50}$ \\
\hline
Prototypical VoteNet~\cite{zhao2022prototypical} &12.39 & 1.52 &	14.54 &	3.05 &	21.51 &	6.13 &	24.78 &	7.17 &	29.95 &	8.16\\
\hline
MetaDet3D\cite{MetaDet} &	6.77 &	0.73 &	8.29 &	1.21 &	15.37 &	2.99 &	19.60 &	4.67 &	24.22 &	5.68 \\
Generalized FS3D\cite{GeneralizedFS3D} & 6.81	& 1.58	& 12.21 & 2.02 &	17.52 &	4.69 &	22.12 &	5.97 &	22.84 &	6.76 \\
P-VAE\cite{PrototypicalVAE} &	14.36 &	2.42 &	22.28 &	4.30 &	27.70 &	8.73 &	\textbf{31.55} & 13.84 & 33.21 &	\textbf{13.98} \\
\hline
\bfseries CP-VoteNet &	\bfseries 16.03 &	\bfseries 4.09 &	\bfseries 22.88 &	\bfseries 6.85 &	\bfseries 28.98 &	\bfseries 11.22 &	30.86 &	\bfseries 13.93 &	\bfseries 34.62 &	13.56 \\
\hline
\end{tabular}
}
\end{table}

\begin{figure}[t]
\centering\includegraphics[width=\textwidth]{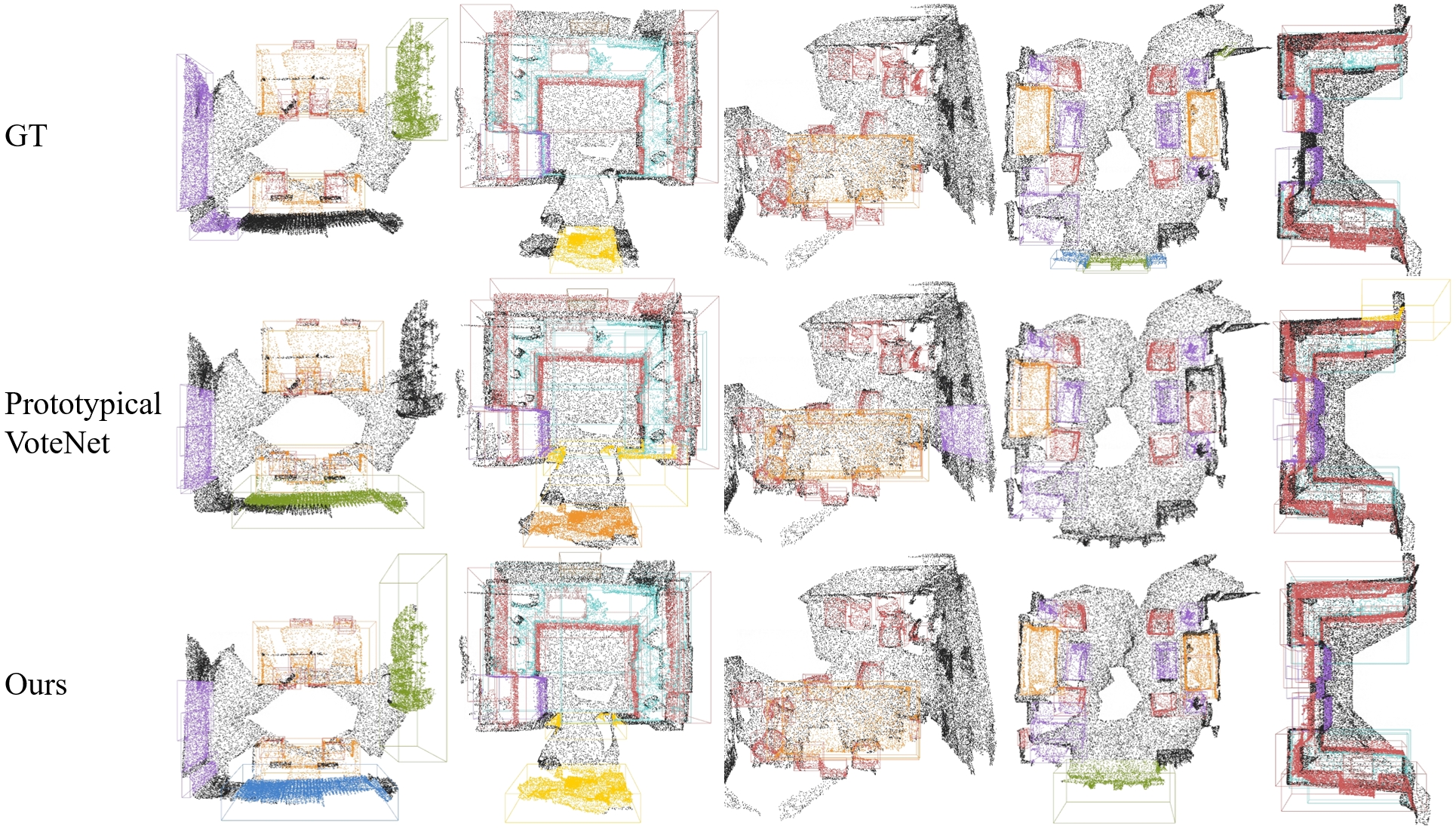}
\caption{Visualisation of few-shot 3D object detection results on point clouds by our method and Prototypical VoteNet \cite{prototypicalFSL} on novel split-1 of \textbf{FS-ScanNet} with $k$=5.} \label{fig:vis}
\end{figure}

\subsection{Results}
Our method is compared with the Prototypical VoteNet~\cite{zhao2022prototypical} baseline and three other FS3D methods including MetaDet3D~\cite{MetaDet}, Generalised FS3D~\cite{GeneralizedFS3D}, and P-VAE~\cite{PrototypicalVAE}.

\subsubsection{Quantitative results}  Tab. \ref{tab:ScanNet} and Tab. \ref{tab:SUNRGBD} present the quantitative results of different methods on FS-ScanNet and FS-SUNRGBD, respectively. As shown in these tables, our CP-VoteNet exhibits a substantial improvement in detection performance compared to the Prototypical VoteNet baseline and, moreover, surpasses the top-performing P-VAE in most task settings. On the FS-ScanNet dataset, The AP$_{25}$ of CP-VoteNet shows an improvement of 2.4 to 4.8 percent over the baseline, while AP$_{50}$ sees an enhancement ranging from 1.1 to 6.5 percent. Compared to the current state-of-the-art, our method also achieves an improvement of up to 5.3 percent. On the FS-SUNRGBD dataset, our method achieves an improvement of 3.6 to 8.3 percent in AP$_{25}$ and 2.5 to 6.7 percent in AP$_{50}$ compared to the baseline. It also demonstrates a marked improvement of up to 2.5 percent in AP$_{25}$ and AP$_{50}$ over the current SOTA P-VAE in the 1-3 shot settings, and matches P-VAE's performance in the 4 and 5 shot settings. Furthermore, given that our positive and negative pairs are constructed within a minibatch, our approach enhances the detection performance of the network without significantly increasing the computational burden and time, thereby maintaining the model's compactness.

\subsubsection{Qualitative results} To qualitatively showcase the superior performance of our approach, we visualise and compare the detection results by CP-VoteNet and Prototypical VoteNet~\cite{zhao2022prototypical} for five point cloud scenes from novel split-1 of FS-ScanNet under the $k$=5 set-up in Fig.~\ref{fig:vis}. Detection boxes of objects of different categories (and the points within them) are marked in different colours. It is evident that the regularisation of semantic features through semantic contrastive learning makes the model's category judgments more accurate, and the constraints imposed on geometric features by primitive contrastive learning enable the network to more precisely predict objects to be detected with corresponding geometric structures, while also reducing false positive boxes from backgrounds or objects of other categories that lack similar geometric information.

\subsection{Ablation Studies}\label{section:ablation}
\subsubsection{Effectiveness of contrastive components} Tab. \ref{tab:ablation} presents the effects of using semantic contrastive learning and primitive contrastive learning on detection performance. This ablation study is conducted using FS-SUNRGBD under the settings of $k$=1, 2, 3, 4 and 5. It can be seen that, under all experimental setups, using SCL or PCL alone results in suboptimal outcomes; the best experimental results are obtained by combining them. This not only proves the effectiveness of two contrastive learning strategies but also confirms our hypothesis: FS3D must take into account both the network's discriminative and generalisation abilities, with the discriminative ability being enhanced by semantic contrastive learning and the generalisation ability being ensured by primitive contrastive learning.

\begin{table}[t]
\centering \scriptsize
\caption{Ablation experiments on the effectiveness of SCL and PCL using \textbf{FS-SUNRGBD}.} \label{tab:ablation}
\begin{tabular}{cc|cc|cc|cc|cc|cc}
\hline
    \multicolumn{2}{c|}{\bfseries Method} & \multicolumn{2}{c|}{\bfseries 1-shot} & \multicolumn{2}{c|}{\bfseries 2-shot} & \multicolumn{2}{c|}{\bfseries 3-shot} & \multicolumn{2}{c|}{\bfseries 4-shot} & \multicolumn{2}{c}{\bfseries 5-shot} \\
    \hline
    SCL & PCL & AP$_{25}$ & AP$_{50}$ & AP$_{25}$ & AP$_{50}$ & AP$_{25}$ & AP$_{50}$ & AP$_{25}$ & AP$_{50}$ & AP$_{25}$ & AP$_{50}$ \\
\hline
\ding{55} &	\ding{55} & 12.39 &	1.52 &	14.54 &	3.05 &	21.51 &	6.13 &	24.78 &	7.17 &	29.95 &	8.16\\
\ding{51} &	\ding{55} & 16.02 & 3.48 & 22.76 & 6.07 & 27.70 & 10.31 & 27.20 & 13.19 & 31.70 & 12.70 \\
\ding{55} & \ding{51} & 15.08 & 2.50 & 21.59 & 4.84 & 26.60 & 10.71 & 29.29 & 12.64 & 33.24 & 13.03 \\
\hline
\ding{51} & \ding{51} &	\bfseries16.03 &	\bfseries4.09 &	\bfseries22.88 &	\bfseries6.85 &	\bfseries28.98 &	\bfseries11.22 &	\bfseries30.86 &	\bfseries13.93 &	\bfseries34.62 &	\bfseries13.56 \\
\hline
\end{tabular}
\end{table}

\begin{table*}[t] \scriptsize
\begin{floatrow}
\capbtabbox{
\begin{tabular}{cc|cc|cc}
\hline
    \multicolumn{2}{c|}{\bfseries Method} & \multicolumn{2}{c|}{\bfseries 3-shot} & \multicolumn{2}{c}{\bfseries 5-shot} \\
    \hline
    $\lambda_1$ & $\lambda_2$ & AP$_{25}$ & AP$_{50}$ & AP$_{25}$ & AP$_{50}$ \\
\hline
0.025 & 0.025 & 28.22 & 11.00 & 31.56 & 10.80 \\
0.1 &	0.1 & \bfseries28.98 &	\bfseries11.22 &	\bfseries34.62 &	\bfseries13.56 \\
0.4 & 0.4 & 25.13 & 8.62 & 31.92 & 9.95 \\
\hline
\end{tabular}
}{
 \caption{Ablation studies on the impact of imbalanced $L_{\text{det}t}$ and contrastive loss using \textbf{FS-SUNRGBD} with $k$=3 and 5.}
 \label{tab:mp}
}
\capbtabbox{
\begin{tabular}{cc|cc|cc}
\hline
    \multicolumn{2}{c|}{\bfseries Method} & \multicolumn{2}{c|}{\bfseries 3-shot} & \multicolumn{2}{c}{\bfseries 5-shot} \\
    \hline
    $\lambda_1$ & $\lambda_2$ & AP$_{25}$ & AP$_{50}$ & AP$_{25}$ & AP$_{50}$ \\
\hline
0.1 & 0.4 & 27.40 & 9.56 & 32.30 & 11.28 \\
0.1 &	0.1 & \bfseries28.98 &	\bfseries11.22 &	\bfseries34.62 &	\bfseries13.56 \\
0.4 & 0.1 &	25.55 & 7.83 & 33.25 & 11.98 \\
\hline
\end{tabular}
}{
 \caption{Ablation studies on the impact of imbalanced $\lambda_1$ and $\lambda_2$ using \textbf{FS-SUNRGBD} with $k$=3 and 5.}
 \label{tab:mp2}
 \small
}
\end{floatrow}
\end{table*}

\subsubsection{Sensitivity of Loss weighting} Tab. \ref{tab:mp} and Tab. \ref{tab:mp2} demonstrate the effect of different values of $\lambda_1$ and $\lambda_2$ on detection performance. To save on computation, this ablation study is conducted on FS-SUNRGBD with $k$=3 and 5. We assign different weights to our contrastive loss to illustrate why we set $\lambda_1$ and $\lambda_2$ to 0.1. As elaborated in section \ref{section:loss}, to achieve the best detection performance, it is necessary to balance the contributions of $L_{\text{det}}$, $L_{\text{semcl}}$, and $L_{\text{primcl}}$ in loss function. An inappropriate proportion of contrastive loss and an imbalance between $L_{\text{semcl}}$ and $L_{\text{primcl}}$ can both impair the model's detection capabilities.

\begin{wraptable}{r}{0.5\textwidth} \scriptsize
    \centering
 \caption{Ablation studies on the effectiveness of projection layers using \textbf{FS-SUNRGBD} with $k$=3 and 5.}
 \label{tab:proj}
 %\vspace{-0.2cm}
    \begin{tabular}{c|cc|cc}
        \hline
        \bfseries Method & \multicolumn{2}{c|}{\bfseries 3-shot} & \multicolumn{2}{c}{\bfseries 5-shot} \\
        \hline
        proj. & AP$_{25}$ & AP$_{50}$ & AP$_{25}$ & AP$_{50}$ \\
        \hline
        \ding{55}  & 24.76  &  8.93 & 31.07 & 13.29 \\ 
        \ding{51} & \bfseries28.98 &	\bfseries11.22 &	\bfseries34.62 &	\bfseries13.56 \\  
        \hline
  %\vspace{-6mm}
    \end{tabular}
\end{wraptable}
\subsubsection{Necessity for projection layers} Tab. \ref{tab:proj} studies the impact of projection layers on model performance. By adding and removing $proj_{\text{s}}(\cdot)$ and $proj_{\text{p}}(\cdot)$, we demonstrate the necessity of using them. It is clearly observable that removing projection layers results in a significant decline in network performance, which confirms that these projection layers can mitigate the loss of feature information by buffering the direct impact of contrastive loss on the original features.

\section{Conclusion}
Existing few-shot point cloud 3D object detection methods extract and utilise various prototypes to facilitate detection of novel class objects, but they lack constraints on these prototypes and also fall short in exploring the intrinsic feature space. In this paper, we design a novel and effective CP-VoteNet to tackle the aforementioned issues. CP-VoteNet comprises two principal components: semantic contrastive learning and primitive contrastive learning. The former enhances the network's discriminative capabilities by imposing constraints on features at the semantic level; the latter improves the network's ability to generalise from base categories to novel categories through constraints on primitive geometric structures. The experimental results prove the effectiveness of our method, and further analysis corroborates the rationality of our design.

%
% ---- Bibliography ----
%
% BibTeX users should specify bibliography style 'splncs04'.
% References will then be sorted and formatted in the correct style.
%
\bibliographystyle{splncs04}
\bibliography{CPVoteNet}
\end{document}